\documentclass{article}
\pdfoutput=1

    \usepackage[preprint, nonatbib]{neurips_2020}

\usepackage[utf8]{inputenc} 
\usepackage[T1]{fontenc}    
\usepackage{hyperref}       
\usepackage{url}            
\usepackage{booktabs}       
\usepackage{amsfonts}       
\usepackage{nicefrac}       
\usepackage{microtype}      
\usepackage{csquotes}
\usepackage{graphicx}
\usepackage[colorinlistoftodos]{todonotes}
\usepackage[frozencache=true,cachedir=mint-cache]{minted}
\usepackage{algorithmic}
\usepackage{subcaption}
 
\graphicspath{{figures/}}

\newtheorem{theorem}{Theorem}
\newtheorem{definition}{Definition}
\newtheorem{proof}{Proof}

\title{Toward Human-Level Artificial Intelligence}

\author{%
  Deokgun Park\\ 
  Department of Computer Science and Engineering\\
  University of Texas at Arlington\\
  Arlington, TX 76019 \\
  \texttt{deokgun.park@uta.edu} \\
}

\begin{document}

\maketitle

\begin{abstract}
In this paper, we  present our research on programming human-level artificial intelligence (HLAI), including 1) a definition of HLAI, 2) an environment to develop and test HLAI, and 3) a cognitive architecture for HLAI. The term AI is used in a broad meaning, and HLAI is not clearly defined. I claim that the essence of Human-Level Intelligence to be the capability to learn from others' experiences via language. The key is that the event described by language has the same effect as if the agent experiences it firsthand for the update of the behavior policy. To develop and test models with such a capability, we are developing a simulated environment called SEDRo. There is a 3D Home, and a mother character takes care of the baby (the learning agent) and teaches languages. The environment provides comparable experiences to that of a human baby from birth to one year. Finally, I propose a cognitive architecture of HLAI called Modulated Heterarchical Prediction Memory (mHPM). In mHPM, there are three components: a universal module that learns to predict the next vector given the sequence of vector signals, a heterarchical network of those modules, and a reward-based modulation of learning. mHPM models the workings of the neocortex but the innate auxiliary units such hippocampus, reward system, instincts, and amygdala play critical roles, too.  
\end{abstract}

\section{Introduction} 

A concrete example of human-level artificial intelligence (HLAI) is a robot that can do many services a human butler could provide. The robot would also converse with other humans and robots to do more tasks. If someone asks for a new dish, it might search the Internet for a recipe and learn to prepare it. Such a robot consists of hardware and software. It looks like a hardware part is almost ready, judging from previous demonstrations, such as Atlas (Boston Dynamics) and Sophia (Hanson Robotics). However, the software part still requires a few more breakthroughs. Therefore, it is important to improve the software part of a butler robot. In this paper, we will introduce our hypotheses on 1) a definition of HLAI, 2) a test of HLAI in a simulated environment where it is administered practically, and 3) a cognitive architecture for HLAI. Then, we will propose a series of experiments to evaluate the  hypotheses.

\section{Laying a common ground by clarifying terms}
The history of AI is long, and recent advances have brought many different flavors to it. Therefore, it would be helpful for our discussion to clarify a few terms such as intelligence, instinct, learning, language, and HLAI. This will explain  why we promote a new term, ~\textit{HLAI}, instead of well-established terms, such as AI or artificial general intelligence (AGI). These definitions draw from an examination of three biological actors - an earthworm, a rabbit, and a human baby --- to distinguish three different levels of intelligence. 

Let us begin with the question of whether an earthworm is intelligent.
The answer will depend on the definition of intelligence. 
Legg and Hutter proposed the following definition for intelligence after considering more than 70 definitions from psychology and computer science~\cite{legg2007universal,legg2007collection}:
\textit{Intelligence} measures an agent's ability to achieve goals in a wide range of environments.

This definition is universal in the sense that it can be applied to a diverse range of agents such as earthworms, rats, humans, and even computer systems. Maximizing gene spreading, or ~\textit{inclusive fitness}, is accepted as the ultimate goal for biological agents~\cite{dawkins2016selfish}. Earthworms have light receptors and vibration sensors. They move according to those sensors to avoid the sun or predators~\cite{darwin1892formation}. These behaviors increase their chance of survival and inclusive fitness~\cite{hamilton1964genetical}. Therefore, we can say that earthworms are intelligent. 
If we agree that an earthworm is intelligent, then we might ask again if it has a \textit{general intelligence}. Considering that it does feed, mate, and avoid predators in a dynamic environment,  it does have general intelligence. However, we would not be so interested in replicating an earthworm-like intelligence. That is why we suggest using HLAI as a term for our community's goal instead of more established terms such as artificial general intelligence (AGI).

Anyways, the moral of the story is that there can be intelligence without \textit{learning}. 
\textit{Behavior policy} is a function that maps a sensory input with the appropriate action. 
The behavior policy of an earthworm is hard-coded and updated only by evolution. In other words, it  is \textit{instinct}~\cite{tinbergen1951study}.
Let us call intelligent animals whose behavior policies are only based on instinct as \textbf{Level 1 intelligence}. The problem with Level 1 intelligence is that evolution is slow to update its behavior policy, and it is costly to convey all variations of behavior policies for diverse environments in the genetic code. 
If an agent can update behavior policy during its lifetime by \textit{learning} new rules such as a new type of food, it would increase the inclusive fitness and reduce the amount of the genetic code for diverse environments. Let's call \textit{experience} as a sequence of sensory inputs (states) and agent actions. A reward can be thought of as a special case of sensory input given by the internal reward system conditioned by the state.  We call those agents with the capability for learning with experience as \textbf{Level 2 intelligence}.


Contrary to our devotion to learning (machine, supervised, unsupervised, reinforcement, self-supervised learning, and so on), most behaviors of Level 2 intelligent agents are not based on learning but are driven by instincts. Let us consider a rabbit that has never seen a wolf before. If the rabbit tries to learn the appropriate behavior by randomly experimenting options when it does encounter a wolf, it is too late to update its behavior policy based on the outcome of random exploration. The rabbit should rely on the instinct which is the Level 1 intelligence. Natural environments are too hostile to use learning as the primary method of building a behavior policy.   

\textbf{Level 3 intelligence} overcomes this limitation by learning from others' experiences.
Bandura pioneered the field with social learning theory~\cite{bandura1977social}. Learning through observation, called observation learning, is found in several species, including non-human primates, invertebrates, birds, rats, and reptiles~\cite{ferrucci2019macaque}. Therefore, Level 3 does not necessarily mean human level of intelligence. However, humans are the epitome of Level 3 intelligence and the only known species using language as a tool for social learning. 
The verbal and written language uses a sequence of abstract symbols to transfer knowledge, relieving the burdensome requirements of observational learning such as presence to demonstrations.  
Thus, we can think of human-level intelligence as \textbf{Level 3 intelligence with language}. 
(Please note that this categorization is the PI's own idea developed  for AI research only: it does no justice to the vast biological diversity. 
For a detailed discussion including limitations, please refer to PI's preliminary work~\cite{mondol2020definition}.)

However, we need to clarify what we mean by learning with language.   Some animals, such as dolphins or monkeys, can communicate with verbal signals. There have been many previous works in the AI field that demonstrated various aspects of language skills. Voice agents, such as Siri or Alexa, can understand the spoken language and can answer simple questions~\cite{kepuska2018next}. AI agents have been trained to follow verbal commands to navigate ~\cite{hermann2017grounded,chaplot2018gated,chen2019touchdown,das2018embodied} and to cook~\cite{shridhar2020alfred}. GPT-3 by Open AI can generate articles published as Op-Ed in The Guardian~\cite{brown2020language,gpt32020}.
Some models can perform multiple tasks in natural language understanding as evaluated in the GLUE benchmark or DecaNLP~\cite{wang2018glue,mccann2018natural}. These  models exhibit performance superior to humans  in all categories except the Winograd Schema Challenge~\cite{levesque2012winograd}, where models perform slightly less than humans  ~\cite{raffel2020exploring}. 
Do these models have HLAI? 

We claim that learning from others' experiences is the essential function of language.   
Let's say you have never tried cola before. Now, for the first time in life, you see this dark, sparkling liquid that may appear ominous. You have a few options for action, including drinking it or running away. Randomly, you select to drink. It tastes good. It rewards you. 
Now your behavior policy has changed so that you will choose to drink it more deliberately next time.  
This is how agents with Level 2 intelligence learn by experience.
Learning with language means that it should bring a similar change in your behavior policy when you hear someone say, ``\textit{Cola is a black, sparkling drink. I drank it, and it tasted good.}''  Based on this observation, we can formally define HLAI as the following. 
\begin{definition}[\textbf{Human-level artificial intelligence (HLAI)}]
An agent has human-level artificial intelligence if  there exists  a sequence of symbols (a symbolic description)   for every feasible experience, such that the agent can update the behavior policy equally, whether it goes through the sequence of sensory inputs and  actions or it receives only the corresponding symbolic description.
\end{definition}

One problem with implementing a test  according to this definition will be to make sure that there exists a symbolic description for every feasible experience  which is not bounded. 

\section{A practical Test for HLAI}

We propose a new test for HLAI, based on the following observation. If a human infant is raised in an environment, such as a jungle, where there are no humans, he or she cannot acquire language. It is \textbf{environment-limited}. Also, if we have animal cubs and try to raise them like a human baby by teaching language, they cannot acquire language. It is \textbf{capability-limited}. Therefore, language acquisition is a function of environment and capability. Based on this argument, we propose the Language Acquisition Test for HLAI as the following; 
\begin{theorem}[\textbf{Language Acquisition Test  (LAT) for HLAI}]
Given a proper environment, if an agent with an empty set of language  can acquire a nonempty set of the language, the agent has the capability for HLAI. 
\end{theorem}

\begin{definition}[\textbf{A set of language}]
A set of language is a set whose element is a tuple of an experience and a symbolic description, where the agent can update behavior policy equally either by going through the experience or by receiving the symbolic description. 
\end{definition}

\begin{proof}
\textbf{Proof by induction} 
Suppose an agent can acquire a new element for the set of language, meaning a symbolic description that can bring the same change for a certain experience without relying on the existing set of language. In that case, the agent can keep adding elements to the set of language for a novel experience until it finds the symbolic description for any given experience.  
\end{proof}


In LAT, a proper environment means that there are other humans to teach language to the learning agent. 
A straightforward way to administer the test would be to  ask human participants to raise the physical robot agent like a human baby. Turing has suggested this approach~\cite{turing1950computing} and the Developmental Robotics community has actively pursued this in several projects~\cite{lungarella2003developmental,asada2009cognitive,cangelosi2015developmental}. However, usage of human participants is cost-prohibitive,  not scalable, and irreproducible. As an anecdote, agents in AlphaStar by DeepMind are said to have experienced up to 200 years of real-time StarCraft game play~\cite{alphastar}. How long would it take for a human to teach language to a simulated baby model?
It would be more useful if we use a simulated environment~\cite{brockman2016openai}.

Indeed, using  simulated environments for grounded language is an active area of research, where agents receive rewards for following verbal instructions in navigation~\cite{chen2019touchdown,savva2019habitat,chaplot2018gated,hermann2017grounded,shridhar2020alfred}  or give correct answers to questions\cite{das2018embodied}. What is remarkable about these works is  the agent can understand the verbal instructions grounded to sensory input thus enabling compositionality of language. For example, let's say that an agent was trained to go to ~\textit{a red small box} and ~\textit{a blue large key} during the training phase.    As a result, during the test phase, the agent can successfully go to a ~\textit{a blue small box}, even though there was no such object during training.

However, the problem is that we don't yet know how we can transfer language skills learned in one domain such as navigation to other domains such as conversation, Q\&A, or cooking.  We conjecture the models learned with explicit rewards are overfitted. More specifically, the action value function of the behavior policy is overfitted to a specific application. This is how Level 2 agents, such as dogs, can learn to follow verbal commands.   Using reward signals generated by environments will be sufficient for the implementation of Level 2 intelligence. However, for Level 3 intelligence, the experiencing reward itself should be part of verbal description. In our previous cola example, there is a part related to the explicit reward that is ~\textit{it tasted good}. In the previous researches, they could teach  the concept of the ~\textit{black sparkling drink} by giving explicit reward when the agent points or navigates to the verbal description. ~\cite{chen2019touchdown,hermann2017grounded,chaplot2018gated,das2018embodied,shridhar2020alfred}. This approach cannot be applied in this case because we require a separate reward mechanism for teaching object concepts,~\textit{black sparkling drink},  and the associated reward,~\textit{it tasted good}.

\begin{figure*}[tb]
  \centering
  \includegraphics[width=0.9\textwidth]{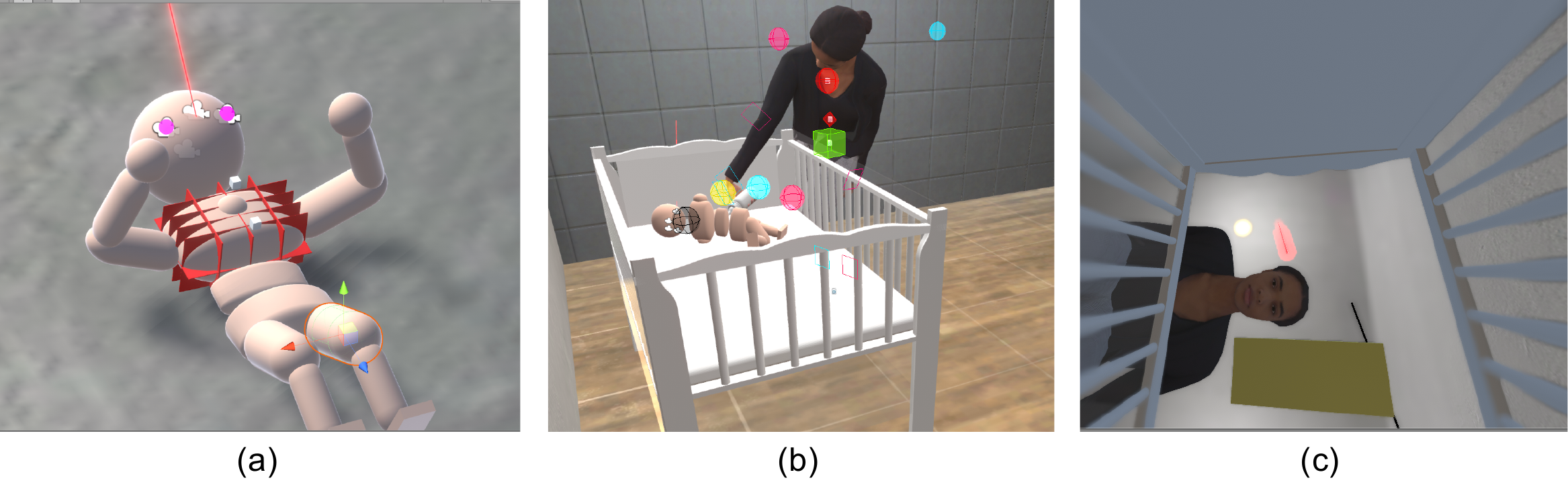}
  \caption{Screenshot of the SEDRo environment. (a) a learning agent which has the physical dimension of the one-year-old human baby. The orange line between eyes represents the eye gaze direction. The grid in the torso shows the area for the distributed touch sensors in the skin. (b) a caregiving agent feeds milk to the learning agent. (c) the visual input to the agent.   }
  \label{fig:sedro}
\end{figure*}

Therefore, we have been working on Simulated Environment for Developmental Robotics (SEDRo) for the practical test of HLAI~\cite{pothula2020sedro}.  SEDRo provides diverse experiences similar to that of human infants from the stage of a fetus to 12 months of age.  
In SEDRo, there is a caregiver character (mother), interactive objects in the home-like environment (e.g., toys, cribs, and walls), and the learning agent (baby). 
The agent will interact with the simulated environment by controlling its body muscles and receiving  the sensor signals according to a physics engine. 
The caregiver character is a virtual agent. It is manually programmed by researchers using a behavior tree that is commonly used in video games to make a game character behave like a human in a limited way. 
Interaction between the agent and the caregiver allows cognitive bootstrapping and social-learning, while interactions between the agent and the surrounding objects are increased gradually as the agent enters more developed stages. 
The caregiver character teaches language by simulating conversation patterns of mothers. SEDRo also simulates developmental psychology experiments to evaluate the progress of intellectual development of non-verbal agents in multiple domains such as vision, motor, and social.   
SEDRo has the following novel features compared to previous works.
  
\begin{itemize}
    \item \textbf{Open-ended tasks without extrinsic reward}  In SEDRo, there is no fixed goal for the agent, and the environment does not provide any reward. Rather than relying on the environment for the rewards, the responsibility of generating rewards belong to the agent itself. In other words, AI researchers have to manually program a reward system to generate reward based on the current state. As an example, if an agent gets a food, the sensory input from stomach will change and  the reward system in the agent will generate a corresponding reward.  

    \item \textbf{Human-like experience with social interaction} Some studies use environments without explicit rewards, and the agents learn with curiosity, or intrinsic reward~\cite{singh2005intrinsically,bellemare2016countBasedExplorationIntrinsicMotivation}.  However, those environments were arbitrary and non-human such as robot arm manipulation tasks or simple games. While such simple environments are effective in unveiling the subset of necessary mechanisms, it is difficult to answer what is a sufficient set.  In SEDRo, we provide a human infant-like experience, because human infants are the only known example of agents capable of developing human-level intelligence. However,  we cannot replicate every aspect of human infants' experience, nor will we try to. There is a subset of experience that is critical for HLAI. Therefore, identifying  these essential experiences  and finding ways to replicate them in the simulation are two fundamental research questions.  Another benefit of a human-like environment is that we can use the experiments from developmental psychology to evaluate the development progress of non-verbal agents.

    \item \textbf{Longitudinal development} SEDRo unfolds agent capabilities according to a curriculum similar to human babies’ development. Many studies suggested that humans or agent models learn faster with constrained capabilities ~\cite{marklee2014longitudinalStudy, keil1981constraintCognitiveDev}. For example, in the first three months, babies are  very near-sighted and do not have any mobility. This makes many visual signals stationary, and the agent can focus on low-level visual skills with eyes. At later stages, when sight and mobility increase, babies can learn advanced skills built-upon lower level skills.  
\end{itemize}

The final benchmark whether the agent has acquired the language will follow the  protocol resembling the previous cola story. 
We give  verbal messages like \textit{``The red ball is delicious(good)''} or \textit{``The blue pyramid is hot (dangerous)''} and check if the behaviour policy toward \textit{the red ball} or \textit{the blue pyramid} has changed accordingly.  Please refer to the PI's preliminary work about the detailed explanation~\cite{pothula2020sedro}.

 \section{A cognitive architecture for HLAI}
 A brain is to generate appropriate behaviors. 
 Biological agents that do not move, do not have brains. 
 The sea squirt has a brain when it is moving in the ocean, but after settlement it does not move and it resorbs most of its brain~\cite{cosmides1997evolutionary}.
 Some behaviors are innate (instinct), and others are acquired (learned).
Therefore, components in the human brain can be divided into universal and specialized parts. The neocortex is a universal module where   functions  can be rewired even after surgeries~\cite{eagleman2020livewired}.
All other parts except neocortex are specialized.

Considering the prominent role of the neocortex, it is easy to regard it as a main unit, and specialized parts as auxiliary supporting unit.  But those specialized parts are the original brain and neocortex later joined as auxiliary supporting unit. An analogy would be CPU and RAM. Historical computers from mechanical calculators to ENIAC did not have a RAM to hold a program and a data.
ENIAC had accumulators that held about 20 numbers and used plugboard wiring as a programming method that took weeks to configure.
Still, ENIAC was a general purpose computer and Turing complete. Modern computers with Von Neuman architecture came later as an additional feature to already general purpose computer. 
However, the modern computer has a vast RAM, and that it is a key contributor to its flexible and universal functionality. 
Neocortex is like RAM. Both are large scale and uniform, and came later as a novel feature for an already-fully-functioning general purpose information processing system.

The PI's approach to programming HLAI is 1) finding the underlying scientific principle behind the universal module or neocortex, and 2) individually engineering the function of  specialized parts.
 
\subsection{A Principle for Universal Module} Let's start with the universal module. 
Unlike RAM which only stores information,  the neocortex implements a functional logic circuit according to its configuration. 
In this sense, the neocortex is more like a field programmable gate arrays (FPGA) device, where it is vastly uniform but the end result of the  configuration is a functional logic circuit. Like FPGA, the neocortex has a lot of uniform universal units  where each unit has a memory and a dedicated controller with  the program for self-configuration. This unit is called a ~\textit{cortical column}~\cite{mountcastle1997columnar}. 
The memory is blank at birth (tabula rasa), but the program in the dedicated controller updates  the  memory content based on the experience, and the content or configuration of the memory becomes a functional logic circuit.

\begin{figure}[bt]
  \begin{subfigure}{0.35\textwidth}
    \includegraphics[width=\textwidth]{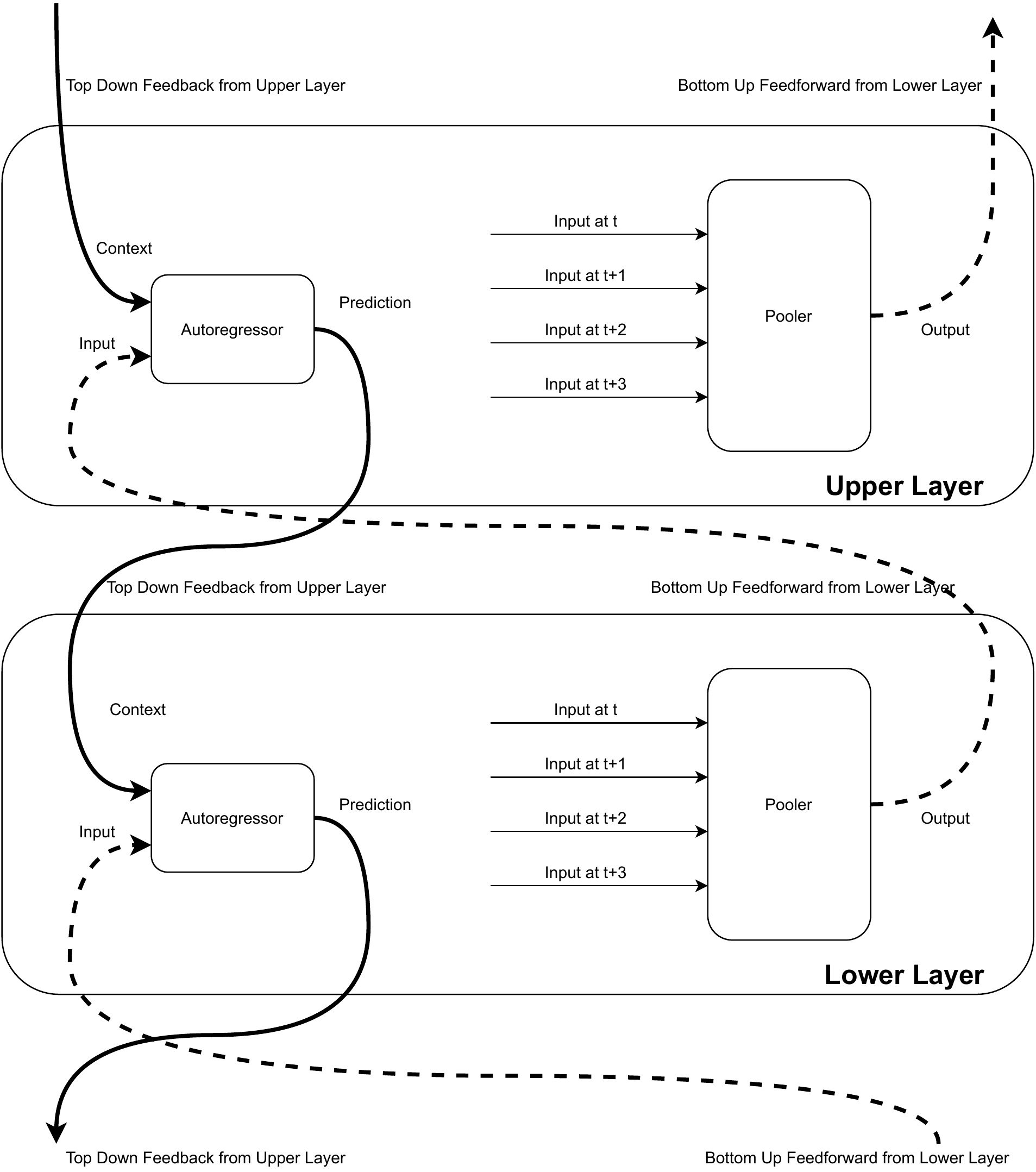}
    
  \end{subfigure}
  \begin{subfigure}{0.64\textwidth}
\begin{minted}{python}
class Layer:
    def __init__(self):
        self.lower = None   # Lower layer
        self.AR = AR()      # Autoregressive model
        self.AE = AE()      # Autoencoder model
        self.pred, self.prevInput = rand(n), rand(n)
    def feed(self, context):
        buffer = []
        for i in range(k):
            self.pred = self.AR.predict(self.prevInput, context)
            input = self.lower.feed(self.pred)
            error = self.pred - input
            AR.update(error)
            buffer.append(input)
        summary = AE.feed(buffer)
        return summary
\end{minted}
  \end{subfigure}%

  \caption{Workings of two layers in Heterarchical Prediction Memory (HPM). The code explains the operation of layers. Each layer receives the input signal from the lower layer and predicts the next vector given the context from the higher layer as a feedback signal. Summary signal will be generated from AE model by summarizing $k$ input signals and be sent to the higher layer as a feedforward input signal.  }
    \label{fig:hpm}
\end{figure}

The PI's hypothesis is that the working principle of cortical columns in neocortex can be explained as  \textbf{modulated heterarchical prediction memory (mHPM)}.
Many conjecture that the main function of a cortical column is to predict the next vector signal given the sequence of vectors that is autoregressive (AR) model~\cite{dai2015semi, hawkins2016neurons, nagai2019predictive, friston2009predictive, rao1999predictive}. Recent advances in language model and image generation with self-supervised learning or semi-supervised learning are based on this~\cite{brown2020language, chen2020generative}.  
Perception is a prediction in the sensory input vector sequence; action is a prediction in the motor command vector sequence. 
Consciousness is a prediction of the vector sequence in the top layer.  
For most verbal thinkers, it is implemented as the prediction in the articulatory rehearsal component (articulatory loop) in the Broca's area commonly referred as~\textit{inner voice}~\cite{buchsbaum2013role, muller2006functional}.

One fundamental issue of the AR model is dealing with long-term dependencies. Hierarchical structures have been used to solve long-term dependencies, such as hierarchical RNN~\cite{chung2016hierarchical, du2015hierarchical, qiu2019neurally} or hierarchical RL~\cite{botvinick2012hierarchical, barto2003recent}. We propose to improve them by two ideas, 1) providing feedback signal from higher layer as context, and 2) doing online and local update instead of end-to-end learning approach. In mHPM, each layer has an AR module which predicts a next vector given input from the lower layer. The prediction is routed to the lower layer as a feedback signal, which provides a context for the AR module in the lower layer.  The lower layer provides a feedforward signal to higher layer that is the summary of $k$ input sequence in the lower layer.  Autoencoder (AE) model is used to make a summary of $k$ inputs. Figure~\ref{fig:hpm} represents this structure when $k = 4$ with the pseudo code. Please note that the lower layer has to run $k$ prediction steps to generate one input for the higher layer. Therefore, temporal frequency and the convergence speed are slower in higher layer which is supported by the biological evidence ~\cite{kiebel2008hierarchy, murray2014hierarchy, runyan2017distinct}.

\begin{figure}[bt]
    \centering
    \includegraphics[width=0.8\textwidth]{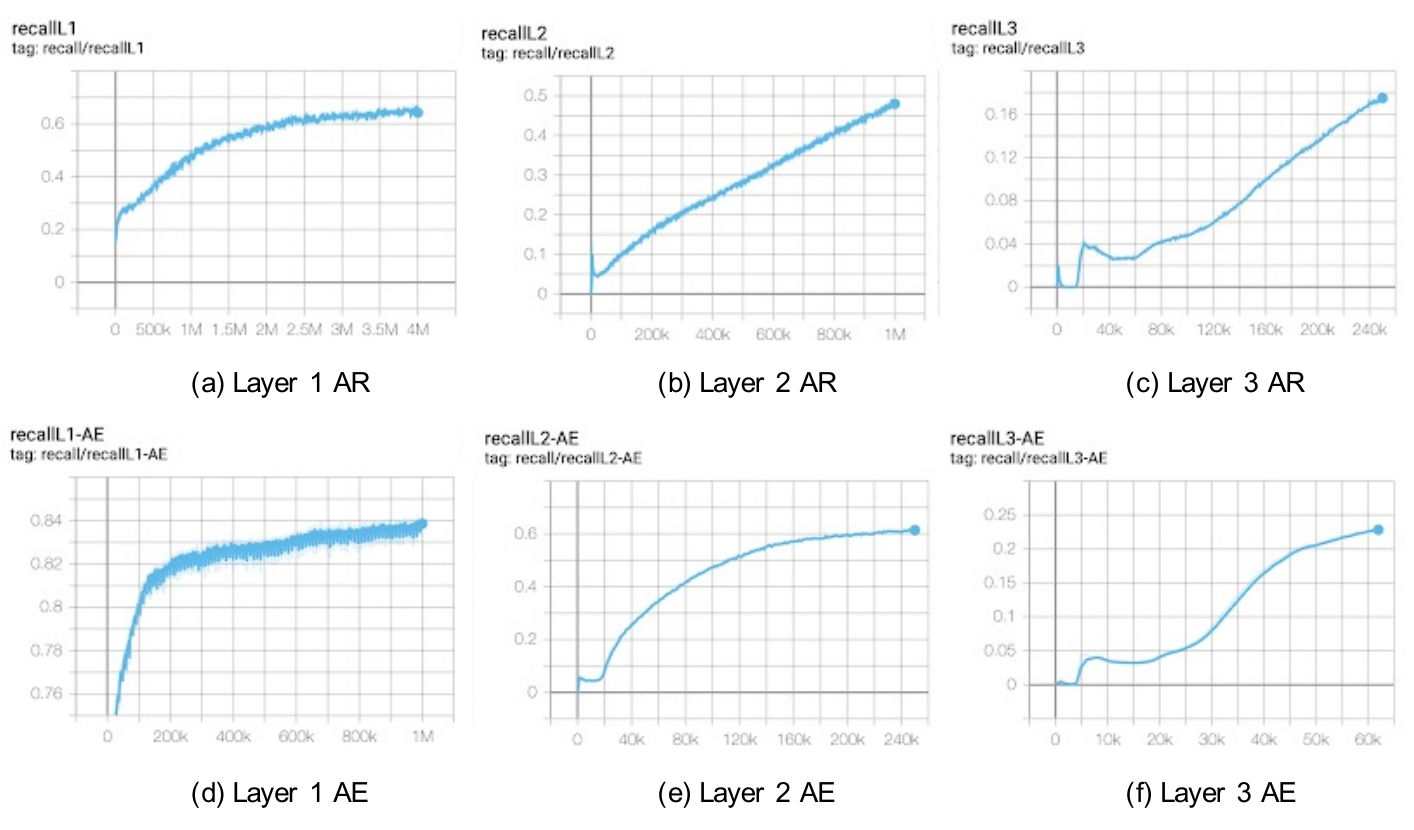}
    \caption{Performance of mHPM for character-level language modeling task with corpus size of 0.5M characters. The chart shows a prediction accuracy for autoregressive (AR) and autoencoder (AE) for layer 1, 2, and 3. (the higher the more accurate). It is interesting that the prediction accuracy of higher layers do increase considering that their input sequence is from the embeddings of an AE model in a lower layer which is also learning from scratch. }
    \label{fig:performance}
\end{figure}

\begin{figure}
    \centering
    \includegraphics[width=1\textwidth]{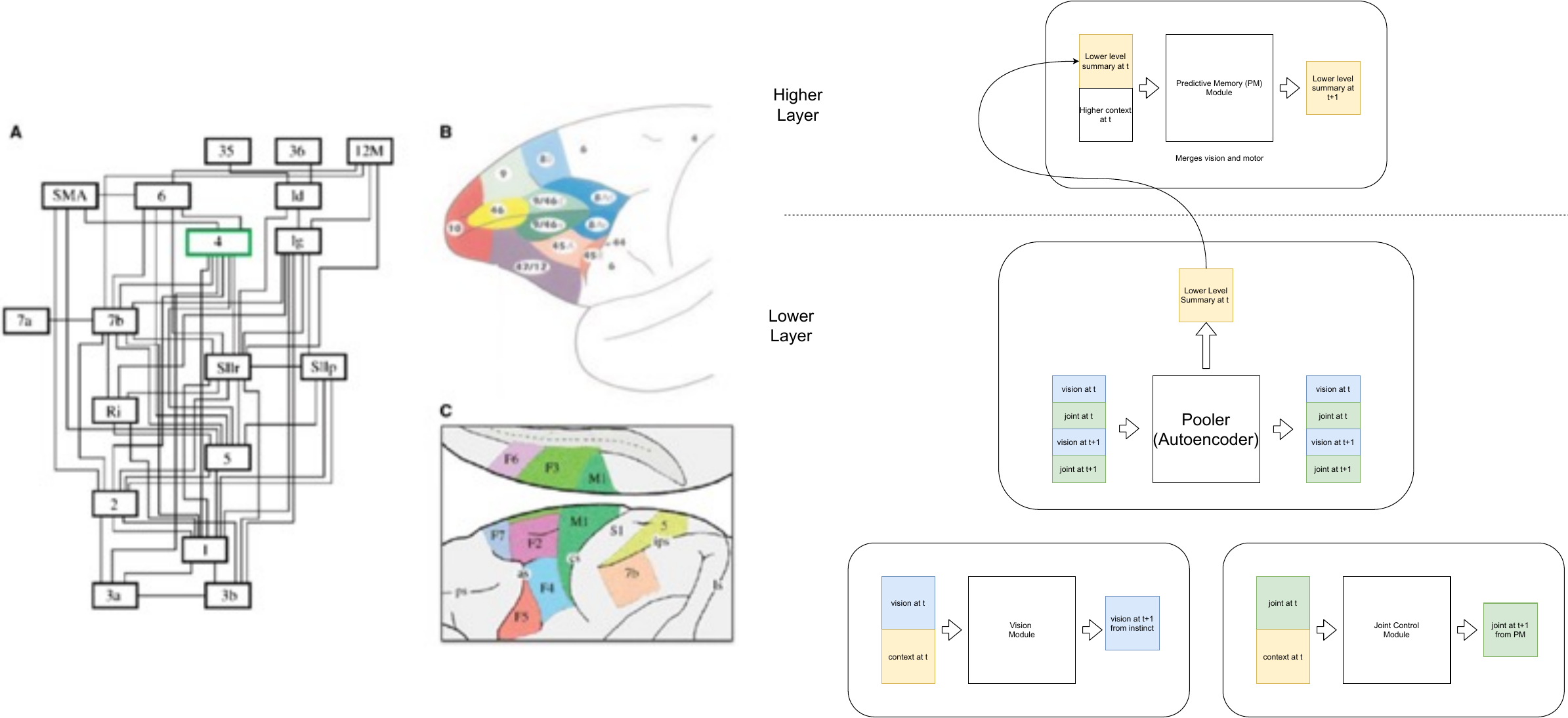}
       \caption{Left: The somato-motor hierarchy for a monkey brain by ~\cite{felleman1991distributed}. Right: Merging Visual pathway and motor pathway by merging signals using autoencoder and feed it forward as input signal for  higher layer. Prediction from the higher layer is feedbacked to the both lower layers (sensory and motor) as the context input. }
       \label{fig:monkey_brain}
\end{figure}

Figure ~\ref{fig:performance} shows the convergence when we used three layers. This was trained as a character-level language model with the training corpus size of 0.5M characters. This shows the preliminary evidence that this structure converges for sequence modeling. However, the previous end-to-end learning models such as attention or recurrent models outperform in language modeling. However, the main benefit of our structure is the AR and AE model in each layer can be updated online and locally.
In end-to-end learning models, the ground-truth for prediction is available only in the bottom layer. Therefore all higher layers are updated with the backpropagation of prediction error of the bottom layer.  In mHPM, all AR and AE models have the local ground-truth in every time step. In this sense, every module in mHPM is self-sufficient for self-supervised learning. This enables the parallel and asynchronous update of individual modules. Therefore, we can build a complex heterarchical network. Figure~\ref{fig:monkey_brain}  shows the somato-motor hierarchy for a monkey brain that is an example of a heterarchical network  ~\cite{felleman1991distributed} (left) and how visual and motor pathways can be combined in mHPM (right).





A heterarchical network is a graph structure with two nodes connected by a link contains an additional information about which node is higher. Compared to a hierarchical network which is commonly represented as tree structure, a node in a heterarchical network can have multiple parents node. For example, reaching a hand to grasp an object requires multiple pathways: the what and where pathway in vision cortex, motor cortex, and somatosensory cortex work together.

However, the mechanism explained so far  simply memorize its sequence regardless of importance or correctness.  In the eyes of a newborn baby, everything will be new. However, there are important sensory inputs, such as the face of the caregiver, and meaningless ones, such as white noise on TV screen. Similarly, a baby moves its arm randomly as motor babbling. But some motor command sequence yields meaningful behavior such as grasping toys, while others do not. Therefore we need a mechanism for learning only meaningful sensory and action sequences. In mHPM, the update rates of AE and AR models are modulated by a reward signal to memorize only meaningful sequences. When there is a reward, the update rate will increase and the sequence will be more likely memorized. The effect of this change is global and lasts for a while (about 2.5 seconds in the human~\cite{yagishita2014critical}).

Below is the  summarization of the characteristics of mHPM;
\begin{itemize}
    \item \textbf{Autoregressive Predictive Memory} Given a sequence of vector signal, the universal module learns to predict the next vector signal. 
    \item \textbf{Heterarchical Network} A lower module feeds information into higher modules in multiple functional domains, while also receiving high-level context signals from them. 
    \item \textbf{Reward-based Modulation} The learning rate is modulated by the global modulation signal generated by the reward system.
\end{itemize}

\subsection{Engineering Specialized Modules for Instincts}

Instinct is an umbrella word for innate behavior policy, and there are different implementation mechanisms including, reflex, emotion, and special-purpose structures. For example, raising arms when tripping, sucking, crawling, and walking are examples of reflex. Reflex relies on dedicated neural circuits. It is useful when it is okay that the response is rigid or fixed  and the reaction duration is instantaneous.  However, when a rabbit hears a wolf cry, the reaction needs to be flexible depending on the context. The reaction state should be maintained over longer time span. Emotion using neurotransmitters or hormones is effective in those cases, because its effect is global, meaning various areas of brain can respond according to it. And it lasts a long while before it is inactivated. Finally, the hippocampus or basal ganglia are special-purpose structures that solve a particular problems such as memory consolidation or decision among conflicting behavior plans~\cite{brown2004laminar}.

However, not all instincts of humans need to be replicated. Of special interests are instincts that enable human level intelligence such as knowledge instinct~\cite{livio2017makes}, social instinct, or language instinct~\cite{pinker2003language}. Below are the instincts that we conjectures as essential and plans to implement it through programming.

\begin{itemize}
    \item \textbf{Minimum reflexes:} Sucking, grasping, crawling, and walking are minimum reflexes that are required to expedite  learning. This can be trained with reinforcement learning, and the resulting behavior policy will be written in the ~\textit{artificial hypothalamus} that will be fixed and not updated during training.  
    \item \textbf{Social instinct:} Innate behaviors such as face recognition, eye contact, following eye gaze, and  attending to caregivers are essential for the social learning. These behaviors are reflexes and can be built using reinforcement learning methods. In addition to those reflexes, a special circuit needs to modulate the reward system  such that a positive voice tone produces a positive reward and vice versa. 
    \item \textbf{Knowledge instinct:} Intrinsic motivation or curiosity plays an important role in the knowledge acquisition~\cite{oudeyer2007intrinsic, singh2005intrinsically, bellemare2016countBasedExplorationIntrinsicMotivation, haber2018learning}. The expected prediction errors in the prefrontal cortex will generate a weak reward in the reward system. 
    \item \textbf{Decision system:} Artificial amygdala will determine the mode of the brain operation among 1) fight or flight, 2) busy without conflicts (Type I), 3) focus (Type II), 4) boring. At the boring state, the knowledge instinct is activated.  Additionally, the ~\textit{artificial basal ganglia} resolves conflict in multiple behavior options with the learning with reward~\cite{brown2004laminar}. 
    \item \textbf{Language instinct:} Social instincts contributes to language acquisition. In addition to it, language specific instincts are babbling, attention to voice-like signals and so on. 
\end{itemize}

\subsection{Summary of the core ideas}
\begin{itemize}
    \item \textbf{Hypothesis I (LAT): } By providing a human-baby-like experience to the learning agent in a simulated environment, we can test the capability for the HLAI. 
    \item \textbf{Hypothesis II (mHPM)}: The universal principle for cortical columns can be explained with modulated heterarchical prediction memory (mHPM).  
    \item \textbf{Hypothesis III (instinct)}: A cognitive architecture, where special-purpose innate modules work with universal learning  modules in non-homogeneous structure, will be more effective than the current structure with only learning modules in homogeneous structure.
\end{itemize} 

\section{Experiment Plan}
 
We explained the prior and preliminary works on 1) a definition, 2) a test, and 3) a cognitive architecture for HLAI in the previous section. There are three  hypotheses,  and we propose a series of experiments to evaluate them in this section. 


The evaluation of Hypothesis I (LAT) is tricky because the environment and the cognitive architecture are dependent on each other. When a learning agent fails to pass the language acquisition test in the simulated environment, it can be either 1) because the environment does not provide sufficient experience, or 2) the cognitive architecture is not capable. The same goes with the Hypothesis II (mHPM) and III (instinct). Therefore, rather than proposing three separate research clusters,  we propose a series of incremental experiments (curriculum) for validation of those hypotheses.  The experiments will start with easy ones and become gradually more difficult in terms of the simulated environments and the  required  mechanisms to solve them. Especially in the early experiments, it  does not depend on the perfect environment that provides a human-like experience such that Hypothesis II (mHPM) can be validated first. Those experiments are formative in the sense solving them requires an engineering solution of both universal module and special-purpose modules. In this sense, the PI's approach is test-driven development. The final test of language acquisition would be summative and conclusive of all three hypotheses.

\subsection{Experiment 1: Integration of Motor and Visual Pathway}
\paragraph{Goal} The goal of experiment 1 is to evaluate Hypothesis II (mHPM) by verifying the feasibility of integrating motor and visual pathway as shown in Figure~\ref{fig:monkey_brain} (right). Eye saccade is an interesting behavior in the sense that it is a simple behavior with three degrees of freedom (DoF), but it involves  many aspects including reflexes, reward-based learning, perception with what and where visual pathways, conflict resolution of innate and learned behaviors, and knowledge instincts. To implement those, motors and the camera sensor of eyes works with lower level cortex including sensory cortex, motor cortex, high-level cortex of pre-frontal cortex, hypothalamus for reflex, reward center with extrinsic and intrinsic rewards, and basal ganglia for conflict resolution. For experiment 1, we will develop the sensory cortex, motor cortex, and merging cortex. 

\paragraph{Protocol}In this experiment we will use a simplified agent mimicking visual saccades.  
The agent has a 3 DoF camera with narrow viewing angle resembling the central vision of the primates.  The agent will be put in SEDRo environment as a natural visual environment. At first, a random vector sequence will be used as a motor control action vectors. In the second part, we will use existing models of  human saccadic eye movement as an input for the motor cortex~\cite{tatler2017latest}.  

\paragraph{Evaluation} We can judge whether the prediction error of the visual input module and the merging module will be decreased over learning. To achieve this, the visual input module has to rely on the feedback connection from the merging module to get the information about the motor command.

\subsection{Experiment 2: Reinforcement learning by the use of a neuromodulator}
\paragraph{Goal} The goal of experiment 2 is to conclude the validity of Hypothesis II (mHPM) by verifying the feasibility of reinforcement learning by using a neurotransmitter. Previous works rely on the estimation of value function for the state to derive the RL. In mHPM, it is done by using a large learning rate when there is a reward, and vice versa.   

\paragraph{Protocol} We will build upon the Experiment 1 setup adding a visually salient moving object as a target resembling a fly. An explicit reward will be generated when the eye focuses on it. 

\paragraph{Evaluation} Evaluation of RL by neuromodulation is to see if moving average of the square error between the object location and the focal point in 3D space is decreasing over time. 

\subsection{Experiment 3: Balancing innate and learned behaviors}

\paragraph{Goal} The goal of experiment 3 is to evaluate Hypothesis III (instinct) by building a decision system that merges innate (instinctive) and acquired (learned) behaviors. Primates have a reflex that foveates to a moving object (pro-saccade) as developed in experiment 2. However, this reflex can be overriden by a training with rewards such that 1) participants maintain the gaze on the center fixation point even though there is a moving object (fixation task), 2) participants maintain the gaze untile the fixation point disappears, and then foveate to the moving object (overlap task), or 3) participants maintain the gaze until the fixation point disappears, and there will be a fixed time interval between the disappearance of fixation point and onset of targets in a fixed location (gap task)~\cite{hikosaka1989functional}. Monkeys can be trained to foveate to moving target (pro saccade) if the fixation point is red, or to move eye in the opposite direction of the target (anti-saccade) if the fixation point is green, too~\cite{everling1999role}.

\paragraph{Protocol} 
We  will use the behavior policy developed in experiment 2 as instinct by fixing it as ~\textit{artificial hypothalamus} and programming ~\textit{artificial basal ganglia} to moderate the two behaviors. (Saccadic reflex is actually hard-coded in the superior colliculus (SC) in the mid-brain~\cite{everling1999role}, while many other instincts are mediated by hypothalamus~\cite{saper2015hypothalamus}.) The reward system will be generated based on the task configuration, such as an anti-saccade task or a fixation task. The reward will modulate the learning rate in the artificial basal ganglia and  the mHPM modules.

\paragraph{Evaluation} The model should  replicate the primate behavior reported in the literature ~\cite{hikosaka1989functional, everling1999role}  

\subsection{Experiment 4: A Few-shot Learning in the Skinner Box}
\paragraph{Goal} The goal is to evaluate Hypothesis III (instinct) by building an ~\textit{artificial hippocampus} for memory consolidation. 
The Skinner Box is an experimental device where a mouse is placed with the switch~\cite{skinner1938behavior}. 
When the mouse push the switch, it  will trigger an electric shock or a food. 
A mouse will learn to avoid or use the switch in a few trials (few-shot learning) compared to much larger number of trials required in reinforcement learning. The hippocampus is commonly accepted as  a mechanism for experience replay that will enable a few-shot learning~\cite{olafsdottir2018role, replay}.   
\paragraph{Protocol} There will be a blue button and red button in the SEDRo simulator.  When the agent presses a button, a positive or negative reward will be delivered. A simplified environment will be used to facilitate interaction with the buttons.
\paragraph{Evaluation} We can examine if the agent acquires behavior policy in a few trials. 

\subsection{Experiment 5: Unity Perception task for Knowledge Instinct}
\begin{figure} 

    \centering
    \includegraphics[width=0.8\textwidth]{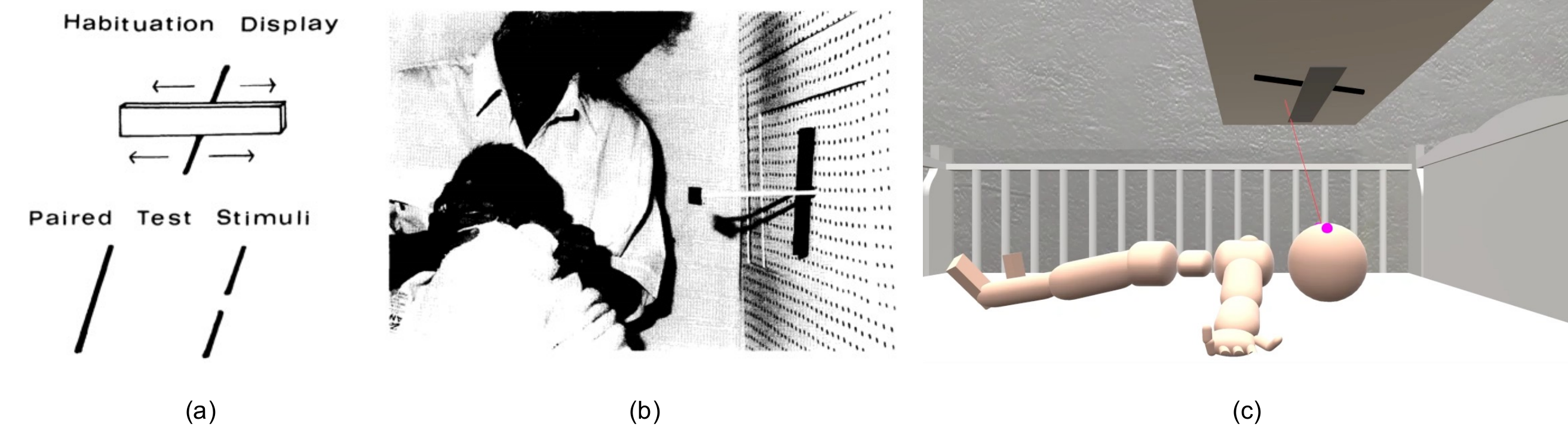}
    \caption{Experiment 5 examines the knowledge instinct. (a) Participants will be shown a rod moving behind the box. After habituation,  the rod will be exposed as a single rod or two rods moving in sync.   Newborn babies till two month will attend to the single rod, while three-months-babies and adults will attend to the two rods, because they attend what is more surprising or unexpected~\cite{slater1990newborn}. (b) The photo of a baby participating in the experiment~\cite{slater1990newborn}. (c) The experiement replicated in the SEDRo~\cite{pothula2020sedro}.}
    \label{fig:unity}

\end{figure}

\paragraph{Goal} The goal is to build  an innate mechanism for curiosity. The prediction error and of the top-most layer and the expected improvement in prediction will be used to generate intrinsic rewards in the reward system which will drive attentive and playful behaviors of agents in a novel situation~\cite{oudeyer2007intrinsic, singh2005intrinsically, bellemare2016countBasedExplorationIntrinsicMotivation, haber2018learning}.  
\paragraph{Protocol} We will replicate the Unity Perception tasks described in ~\cite{slater1990newborn}. Please refer to Figure~\ref{fig:unity} for more explanation about the protocol. There are other experiments for evaluation of non-verbal agents listed in PI's preliminary publication~\cite{pothula2020sedro}.

\paragraph{Evaluation} We can check whether the model recognizes the unity of object by examining which setup the model attends more. 



\subsection{Experiment 6: Language Acquisition Test in SEDRo}
\paragraph{Goal} The goal is to check if language can be acquired in a naturalistic environment. To enable this, it is expected to program a few instincts in social and language domain listed in 1.3.2. This experiment will conclude the validity the Hypothesis I (LAT) and III (instinct). 
\paragraph{Protocol} As explained in 1.2, the agent will be put in a simulated home with a caregiver.  
\paragraph{Evaluation} We can verify if the agent has acquired the language by comparing the change of behavior policy between direct experience of touching ~\textit{red ball} and getting a pain, and receiving verbal description such as ``The red ball is hot (dangerous).'' We will build a set of cases with simple words that can be learned in SEDRo.

\section{Intellectual Merit}
The pursuit of flying machines and thinking machines have many things in common. Both have been long-standing dreams of humankind inspired by the evidence from nature, namely a bird and a brain. While the former has been achieved, the latter remains one of the biggest technical challenges.  One lesson we can learn from the invention of flying machines is that we should be careful when we copy somebody else's answer. You can copy too much or too little. For example, it is still challenging for us to build a drone that flies as birds do. The first flight would have been much delayed if we tried to copy the birds' exact mechanism. It is the case when we try to copy too much. In other extremes, we were not successful when we ignored the biological implementation and worked on creative approaches  such as a flying carpet or hot air balloons. The success came when we found a balance between  too much and  too little.  

In AI research, we insist that we are currently slightly in ~\textit{too little} side of the pendulum. If the proposed work makes a progress, it would provide two generalizable lessons to the community. 
First, current state of the art (SOTA) models tend to rely on end-to-end learning with backpropagation of loss from a single objective function. 
This is difficult to extend to the parallel multitask required in biological agents. We can drive while talking on the phone. We propose to use local and online optimization of modules with self-sufficient self-supervised learning. It will enable heterarchical networks of modules.   Second generalizable lesson for the community would be that we should add non-homogeneous special-purpose modules to the cognitive architecture for an organic mix of innate and learned behaviors. Current SOTA tends to be more homogeneous in its structure emphasizing learning only. Again, contrary to our devotion to various forms of learning, most behaviors are based on instincts. Remember the poor rabbit in front of the wolf in the introduction. Even for human-level intelligence, instincts still play a role for   knowledge and language acquisition. Important questions are ``What instincts enable social interaction, knowledge learning, and language acquisition?'' and ``How do those instincts work?''  Again, this idea is not new, but there have been many cognitive architectures including SOAR and ACT-R. It is because of the recent fantastic success of the Deep Learning approach that the pendulum has swung toward the learning side. It might be worthwhile to consider to balance the pendulum again by incorporating innate mechanisms with learning if the community feels the progress is slowing down. 
If the community try those two ideas, they will find more useful cognitive architectures and advance the AI field significantly.  That will be the main take-away messages for readers.

\bibliography{references}
\bibliographystyle{unsrt}

\end{document}